\RequirePackage{fix-cm}
\documentclass[smallextended]{svjour3}       
\smartqed  
\usepackage{graphicx}
\usepackage{multirow}
\usepackage{rotating}
\usepackage{subfig}
\usepackage{times}
\usepackage{amsmath}
\usepackage{amssymb}
\usepackage{algorithm}
\usepackage{algorithmic}
\usepackage{array}
\usepackage{lineno, hyperref}
\usepackage{bm}
\usepackage{subfig}
\usepackage{enumerate}
\usepackage{float}
\usepackage{color}
\usepackage{marvosym}

\begin{document}

\title{Road Segmentation with Image-LiDAR Data Fusion}

\author{Huafeng Liu \Letter\and
        Yazhou Yao \and
        Zeren Sun \and
        Xiangrui Li \and
        Ke Jia \and
        Zhenming Tang 
}

\institute{
Huafeng Liu \at School of Computer Science and Engineering, Nanjing University of Science and Technology, Nanjing City 210094, China.\\
\email{liu.hua.feng@outlook.com}
\and
Yazhou Yao, Zeren Sun, Xiangrui Li \at School of Computer Science and Engineering, Nanjing University of Science and Technology, Nanjing City 210094, China.\\
\and
Zhenmin Tang \at School of Computer Science and Engineering, Nanjing University of Science and Technology, Nanjing City 210094, China.\\
\email{tzm.cs@njust.edu.cn}
\and
Ke Jia \at School of Computer Science, Chengdu University of Information Technology, Chengdu City, China\\
\email{jiake@cuit.edu.cn}
}

\date{Received: date / Accepted: date}

\maketitle

\begin{abstract}
Robust road segmentation is a key challenge in self-driving research. Though many image based methods have been studied and high performances in dataset evaluations have been reported, developing robust and reliable road segmentation is still a major challenge. Data fusion across different sensors to improve the performance of road segmentation is widely considered an important and irreplaceable solution. In this paper, we propose a novel structure to fuse image and LiDAR point cloud in an end-to-end semantic segmentation network, in which the fusion is performed at decoder stage instead of at, more commonly, encoder stage. During fusion, we improve the multi-scale LiDAR map generation to increase the precision of multi-scale LiDAR map by introducing pyramid projection method. Additionally, we adapted the multi-path refinement network with our fusion strategy and improve the road prediction compared with transpose convolution with skip layers. Our approach has been tested on KITTI ROAD dataset and have a competitive performance. 
\end{abstract}

\section{Introduction}

With the booming of intelligent transportation system research, autonomous driving technology has gained more and more attentions. Road segmentation, as one of the crucial tasks, is a basic topic for enabling autonomous ability and mobility\cite{deepdrive,treml2016speeding}. In road segmentation research, various methods have been proposed to find road area in RGB image\cite{RBNet} or 3D LiDAR point cloud\cite{Luca2017,Chen2017}. However, the colors, textures and shapes can be very different due to the various illumination condition, weather condition and very different scenes, eventually makes road segmentation still a challenging task.

Deep learning is a powerful tool on learning representation in basic multimedia tasks, such as image classification\cite{yao2018tip,yao2018ijcai,tkde2019,tmm2018}, image searching\cite{ijcai2019,shen1,shen2,shen3}, image segmentation\cite{Long2015Fully},  as well as scene recognition\cite{YYZ2018,YYZ2019,XGS2017tcsvt,XGS2019,XGS2017ijcv}. As a common sense, the features used in those tasks have a great impact on final performance, and recently Convolutional Neural Network(CNN) has been demonstrated that, automatic feature learning on massive annotated data surpasses hand-crafted features in many applications. As a result, more and more researchers are trying to exploit Deep Neural Network(DNN) in many fields. Deep convolutional neural networks ,like VGG\cite{VGG} and Residual Net\cite{RESNET}, are used as encoders, acting like de facto standard feature generators in many applications, and they greatly improve the results of all the tasks we mentioned above. Specifically,  in the past 3 years, for the road segmentation task, which is one of the semantic segmentation problems, the performance have been improved dramatically by methods based on the variations of Fully Convolutional Network(FCN)\cite{Long2015Fully}. FCN established a classic encoder-decoder pattern for segmentation using a deep CNN, and this leads to an automatic end-to-end feature extraction and segmentation architecture, which has a giant parameter space to represent diverse objects in a very complex way, thus make them much more classifiable. Also, the deconvolution layers which are widely used used in neural networks\cite{ZMM2017} are also introduced to tackle up-sampling and rebuild the pixel level label prediction. Then, many effective semantic segmentation network were proposed and there has been an amazing performance improvement.

But autonomous driving cars are equipped with various sensors to enhance their environmental perception ability. Though it has been reported that image based road segmentation claims high performance in many tests, road segmentation via single sensor is not that robust in complex scenes. Those above motivate us to develop road segmentation solutions under fusion strategies. 

Our work follows the encoder-decoder pattern however some drawbacks have to be discussed firstly:

1) Although there are a lot of successful works and public datasets on pure image based road segmentation, their weaknesses are obvious: they are insufficient to learn a robust representation of road area, due to the lack of sample quantity and scene diversity across datasets. At the same time, learning 3D geometric information is difficult since recovering 3D structure from 2D image remains a challenging problem nowadays. To address this issue, LiDAR based methods\cite{Chen2017,Luca2017} have been proposed these years. However sparse LiDAR point cloud is not always helpful to improve segmentation performance because sparse LiDAR point is location accurate but visual perception deficient. Therefore, fusing multi-sensor inputs to improve road segmentation performance as well as maximize sensor utilization is an intuitive idea\cite{Han2016,Schlosser,asvadi}. In the fusion pipeline a alignment procedure is usually required. Before feeding image and LiDAR into processing system, they are supposed to align with each other via calibration parameters. Then, spatial and geometric features embedded in LiDAR point cloud can be extracted simultaneously with image features via an end-to-end deep neural network.  However, since the design of network structure is very flexible and there are various deformations to fuse data, how to exploit image and LiDAR information within the CNN based semantic segmentation network remains an open problem. 

\begin{figure}[h]
	\centering
	\includegraphics[width=\textwidth]{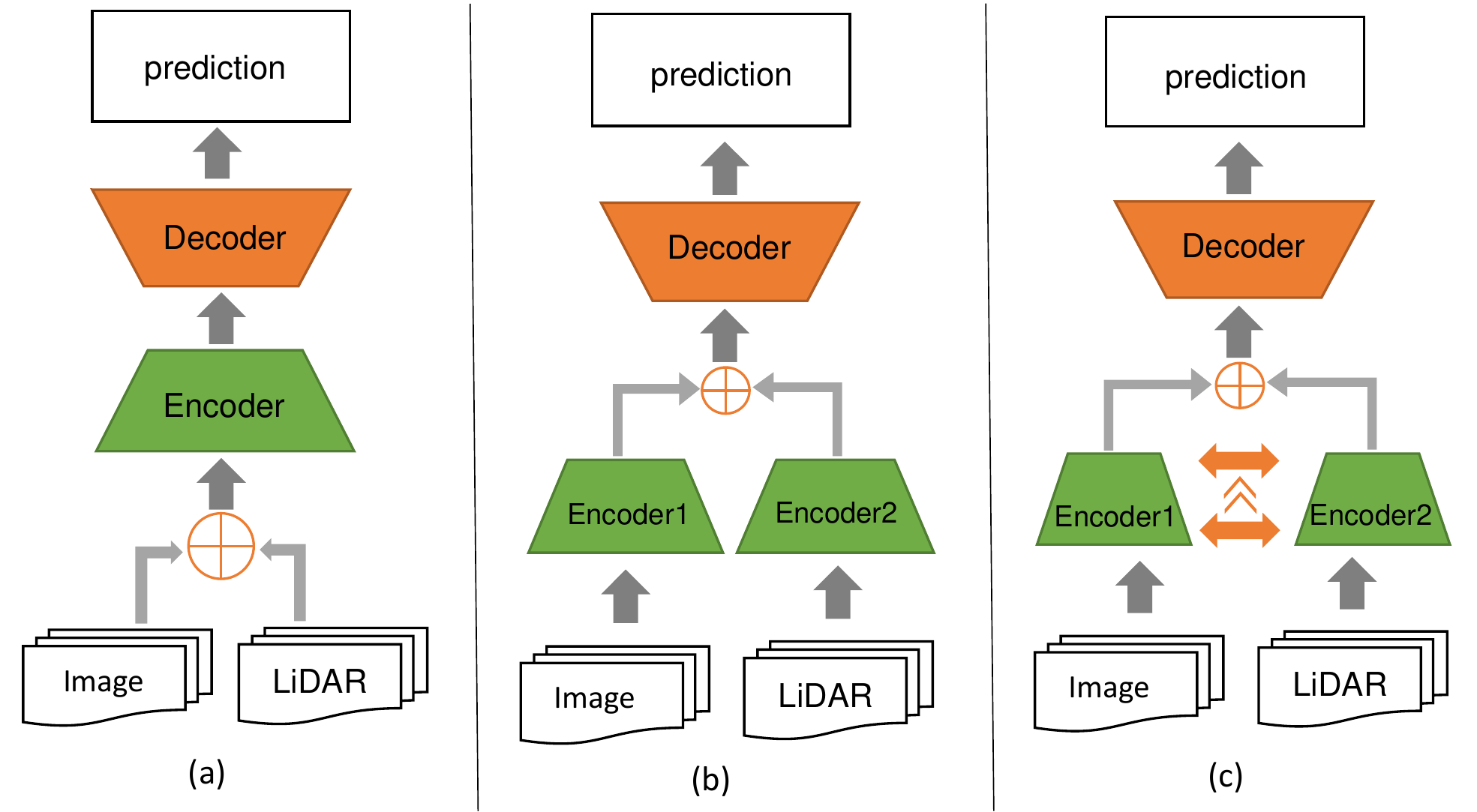}
	\caption{The three typical fusion structures at encoder side. Figure(a) depicts  early fusion mode, in which image and LiDAR are fused at the input layer. Figure(b) is late fusion in which two side-by-side encoder will merge at the end of encoding stage(before decoding). Figure(c) is stage-wise fusion, meaning that fusion happens at every stage inside the encoder.}
	\label{fig:three_fusion_mode}
\end{figure}

2) With the developing of deep learning related research, the encoder and decoder of original FCN is not good enough since we have more computing resources so that we can endure a much deeper network. Besides, as we mentioned above, in this novel method, the network model must have the ability to fuse data from different kinds of sensors. Many existed methods embed their fusion structure in encoder stage, expecting fuse information by synergetic  feature extracting steps\cite{LiDARCam}, or, just after a series of side-by-side encoding procedures\cite{Schlosser}. In addition, stage-wise fusion like cross-fusion\cite{LiDARCam} and siamese-fusion\cite{siamese} have been reported recently. The above mentioned fusion mode is illustrated in Figure.\ref{fig:three_fusion_mode}. Generally, those attempts improved the performance by fusion features in the encoder step, as they reported, but makes pre-trained image encoder hardly useful. As a consequence, we need a very big dataset to train a pre-trained fusion model as a alternative to maintain network performance, which is not realistic.  Therefore, in our work, the encoder is updated to residual net and the decoder is replaced by a more complex model, in which data from different sensors are fused in multiple scales to achieve a better prediction result. 

Motivated by tackling drawbacks discussed above, we propose a novel approach with the following contributions: 

1. We tried a novel fusion structure design in which the LiDAR fusion is performed in decoder instead of encoder. This designation makes it possible to utilize pre-trained models in encoder easily, also, generate better label prediction by enhancing the up-sampling.

2. We use pyramid multi-scale re-projection instead of classic step-by-step pooling method to generate multi-scale LiDAR map for fusion in each stage, which alleviate degeneration in down-sampling . 

3. A compact fusion and up-sampling structure is designed to perform segmentation prediction. We conduct a series of experiments of KITTI ROAD dataset\cite{KITTI} and it demonstrate that the performance of our method is competitive to recent state-of-the-art ones. 

\section{Related Works}
Computer vision and pattern recognition research cover a lot of fields, and the a an important topic is feature extraction and analysis\cite{ywk,ywk2,zwm,ywk3,mmm,aim,huangpu}. In the past decade, deep learning with CNN becomes a very important feature extractor for classification and segmentation problems \cite{neurocom,prl,acmmm}. Since road segmentation is a semantic segmentation task, the following subsection is the traditional ideas on road segmentation. Then the subsequent subsection reviews deep neural network based semantic segmentation as our work are inspired by many recent progress in new architecture in CNN classifier and semantic segmentation methods. Finally, data fusion methods in previous works is listed.

\textbf{Road segmentation}. Before the deep learning comes into vogue, the traditional methods were studied under probabilistic framework. Amongst these methods, the most popular idea is extracting hand-crafted features to perform a pixel-wise prediction, finally mark the road and non-road area. For example, Keyu Lu et al. proposed a hierarchical approach for road detection\cite{LUK2014}. They trained a Gaussian mixture model (GMM) to obtain road probability density map(RPDM), then divided the images into superpixels. They tried to select road superpixels from seeds in a growcut framework firstly, and refined the results with a conditional random field (CRF). Liang Chen et al. only use Lidar point clouds to detect the road\cite{Chen2017}. They resampled the Lidar point clouds and generated Lidar-imageries, and proposed a Lidar-hisotgram derived from them. In the Lidar-histogram representation, the 3D traversable road plane in front of vehicle can be projected as a straight line, and the positive and negative obstacles are projected above and below the line respectively. Liang Xiao et al. proposed a hybrid CRF to fuse Lidar and image data\cite{Xiao2018}. They firstly extracted features from images and used an boosted decision tree classifier to predict the unary potential. Then the pairwise potential was designed via hybrid model of the contextual consistency in the images and Lidar point clouds, as well as the cross consistency between them. Finally they used this CRF to detect the road areas. Those method contains an Achilles' heel: hand-crafted features is to difficult to produce but semantic information requires massive features and their efficient combination. To strengthen the weaknesses mentioned above, the deep methods  are introduced to this field.

\textbf{Semantic Segmentation}. FCN is the first well-known method for end-to-end deep semantic segmentation\cite{Long2015Fully}. FCN's designation follows the encoder-decoder pattern with transposed convolutions and skip layers, this architecture laid the foundation for segmentations. In the meanwhile, SegNet\cite{segnet} use max-pooling indices in the decoders to perform upsampling of low resolution feature maps which retains high frequency details in the segmented images as well as reduces the total number of trainable parameters in the decoders. To make strong use of data augmentation of available annotated samples, UNet\cite{unet} develops architecture consists of a contracting path to capture context and a symmetric expanding path that enables precise localization. DeepLab series now evolve to DeepLabV3+\cite{deeplab}, it extends DeepLabv3 by adding a simple yet effective decoder module to refine the segmentation results especially along object boundaries. Some work achieve very good result by combining FCN with Conditional Random Field(CRF) using recurrent net\cite{sisdcncrf2015,CRFasRNN}.
\begin{figure}[h]
	\centering
	\includegraphics[width=0.7\textwidth]{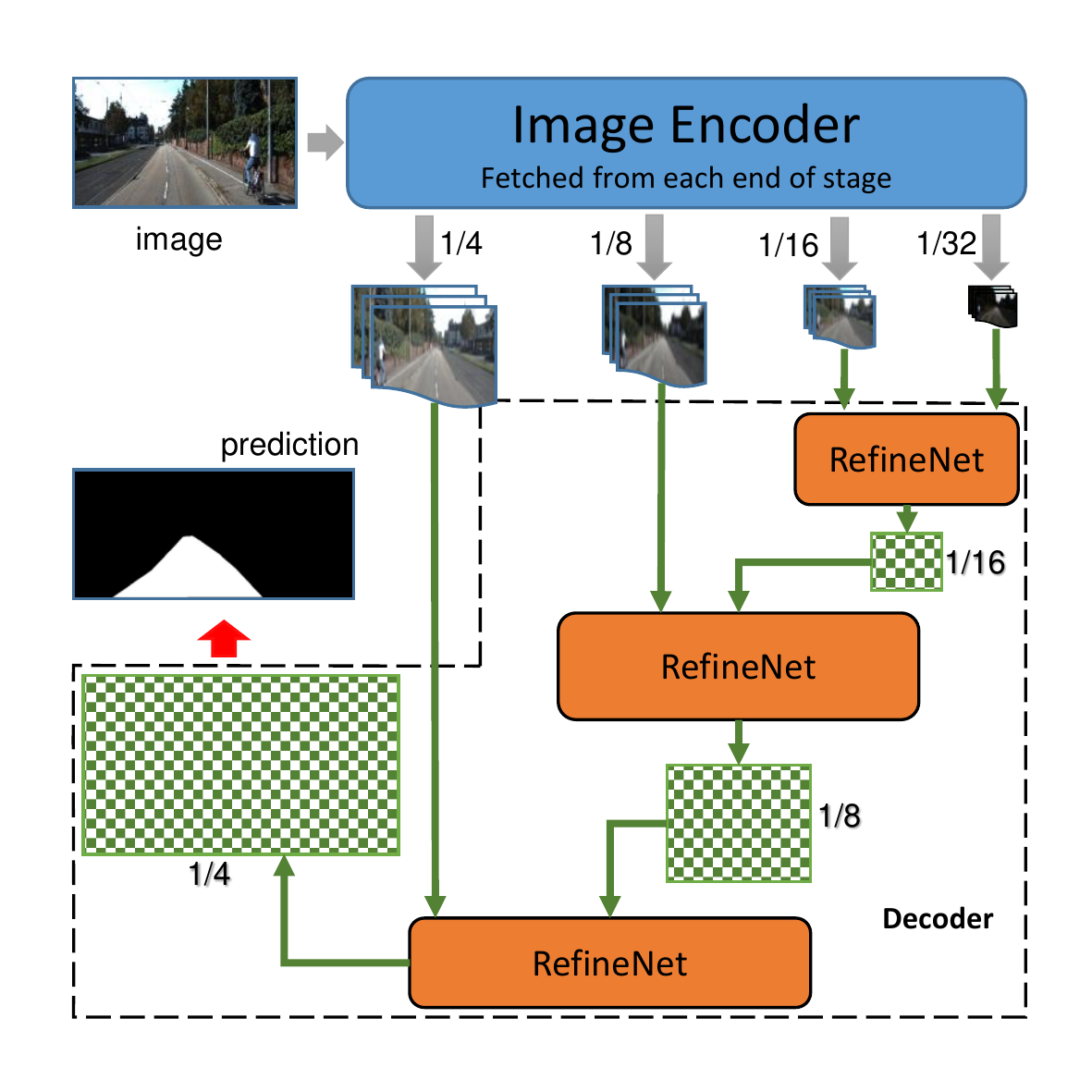}
	\caption{Architecture and work flow of RefineNet. This picture depicts the reference work flow of 3-cascaded $RefineNet$. Some unnecessary implementation details are omitted in both two pictures  just for brevity.}
	\label{fig:refinenet}
\end{figure}

\textbf{RefineNet}. Guosheng Lin present a generic multi-path refinement network called RefineNet. RefineNet is a generic multi-path refinement network in which the information available along the down-sampling process are explicitly exploited to enable high-resolution prediction using long range residual connections\cite{refinenet}. The most important component in RefineNet is Chained residual pooling(CRP). The key idea of RefineNet\cite{refinenet} is deeper layers that capture high-level semantic features can be directly refined using fine-grained features from earlier convolutions. The experiments have proved that it achieved state-of-the-art performance in many semantic segmentation datasets. This network is elastic since there exists many configuration after encoder. Our work is based on the principle of RefineNet and a fusion module is added to CRP module. Figure. \ref{fig:refinenet} is an illustration of RefineNet.

\textbf{Data Fusion} for DNN based road segmentation. Schlosser et al. discussed different fusion strategies, named early fusion and late fusion\cite{Schlosser}. Last year, Caltagirone et al. proposed a new fusion architecture name corss-fusion to directly learn from data where to integrate information by using trainable cross connections between the LIDAR and the camera processing branches\cite{LiDARCam}. A siamese fusion method for road segmentation is also proposed to fuse LiDAR and camera informations\cite{siamese}. On road segmentation with camera and LiDAR fusion tasks, the only suitable dataset and benchmark, to our knowledge, is KITTI ROAD dataset\cite{KITTI}. This dataset provide various sensor data collected on a real car with careful calibration. Due to above limitation, we will test our method using KITTI benchmark.

\section{Data Fusion for Road segmentation}

\subsection{Architecture and Work Flow}
\begin{figure}[h]
	\centering
	\includegraphics[width=\textwidth]{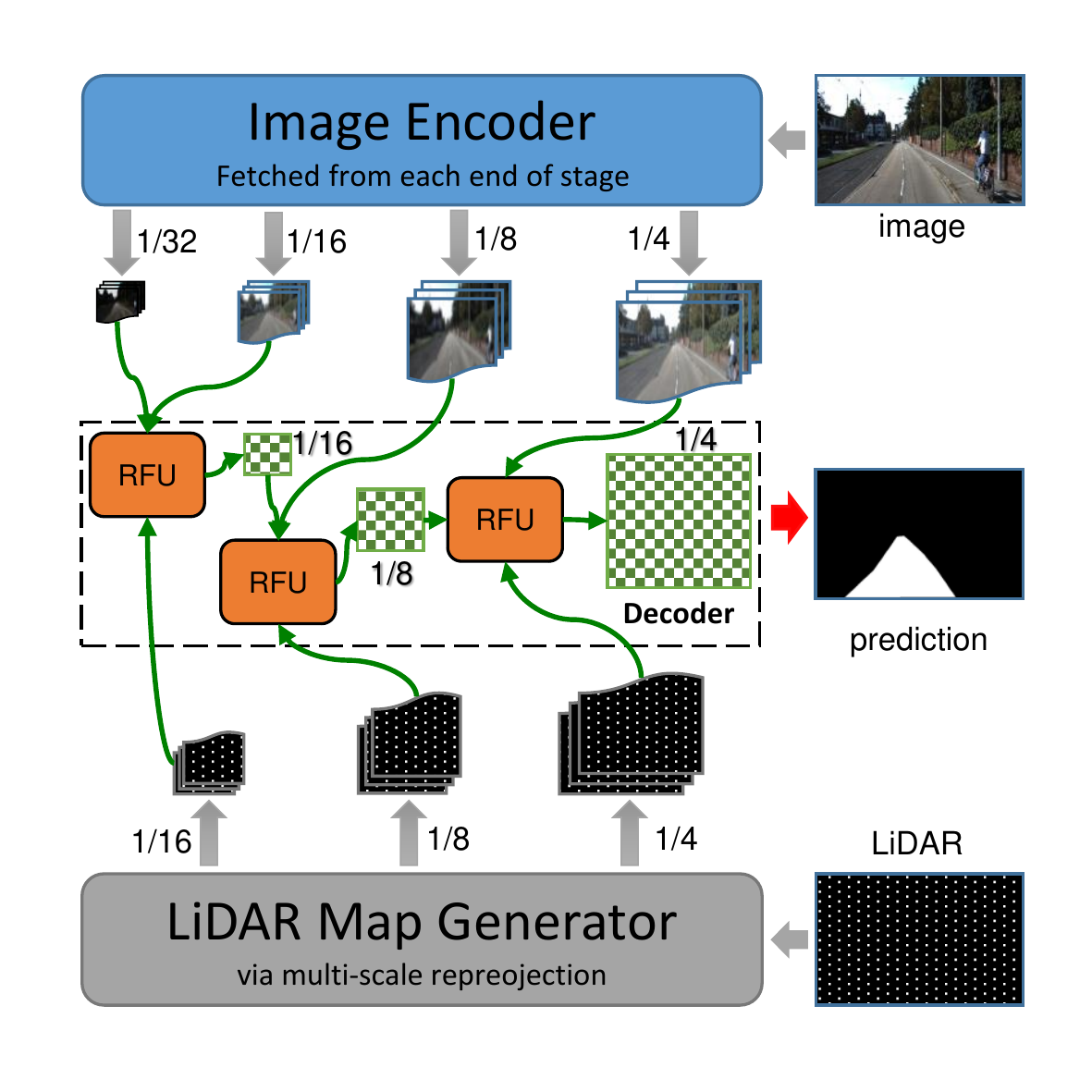}
	\caption{Architecture and work flow of our proposed network. The orange rectangle labelled $RFU$ is short for Refined Fusion Unite, in which LiDAR information fusion and up-sampling is processed. Some unnecessary implementation details are omitted in both two pictures just for brevity.}
	\label{fig:architecture}
\end{figure}

Our approach follows the encoder-decoder design pattern. Encoder module is in charge of generating feature maps at different scales for image input. Decoder module predict the segmentation result by gradually up-sample the feature maps while fusing LiDAR information. The summery of our approach is depicted in left side of Figure. \ref{fig:architecture}

\textbf{Encoder}. The encoder as backbone in our network is $ResNet50$ which down-sample the image to $1/32$ and expands the feature channels from $3$ to $2048$. In fact, $VGG16$, $ResNet101$ is also suitable as an encoder in our task. The key point of encoder is that multi-scale feature maps is required during the down-sampling procedure. Taking different structures of encoder into consideration, we utilize the consecutive down-sampled feature maps $S_i$ form $1/4$ to $1/32$ of original scale, denote as $S_i \in [\frac{1}{4}I, \frac{1}{8}I,\frac{1}{16}I, \frac{1}{32}I]$. Each $S_i$ will be used to refine the up-sampling prediction. In our network, we use  pretrained model on ImageNet\cite{imagenet} in encoder stage, since encoder is only in charge of image feature map generation.

\textbf{Decoder}. Our decoder is inspired by multi-path refinement networks\cite{refinenet} which is optimized for high resolution image segmentation. We make a major revision to RefineNet block by embedding a fusion step to process LiDAR information. The Original RefineNet blocks designed two additional modules named residual convolutional unit(RCU) and chained residual pooling(CRP). In RefineNet, RCU and CRP borrows the idea from residual blocks in which gradients can be directly propagated, thus reduced the training difficulty. We discard the RCU and replace it with LiDAR processing block to generate LiDAR feature map and fuse it with image information. Additionally, CRP module has been slightly modified to reduce the parameter numbers. Our design is aiming at embedding a LiDAR fusion module without enlarge the network too much.

\textbf{Fusion}. There are two fusion strategies inside our architecture. One fusion is performed among image feature maps at multiple scales, whose objective is refining image up-sampling with more details. This technique has been proven effective in many semantic segmentation networks, such as skip layers of FCN and similar structure in Unet. The other one happens between LiDAR feature maps and image feature maps. We designed this sensor level fusion structure to utilize image and LiDAR information at same scale. The details of fusion will be discussed in Section. \ref{sect:fusion}

\subsection{Fuse Image and LiDAR information in Decoder}
\label{sect:fusion}

\begin{figure}[h]
	\centering
	\includegraphics[width=\textwidth]{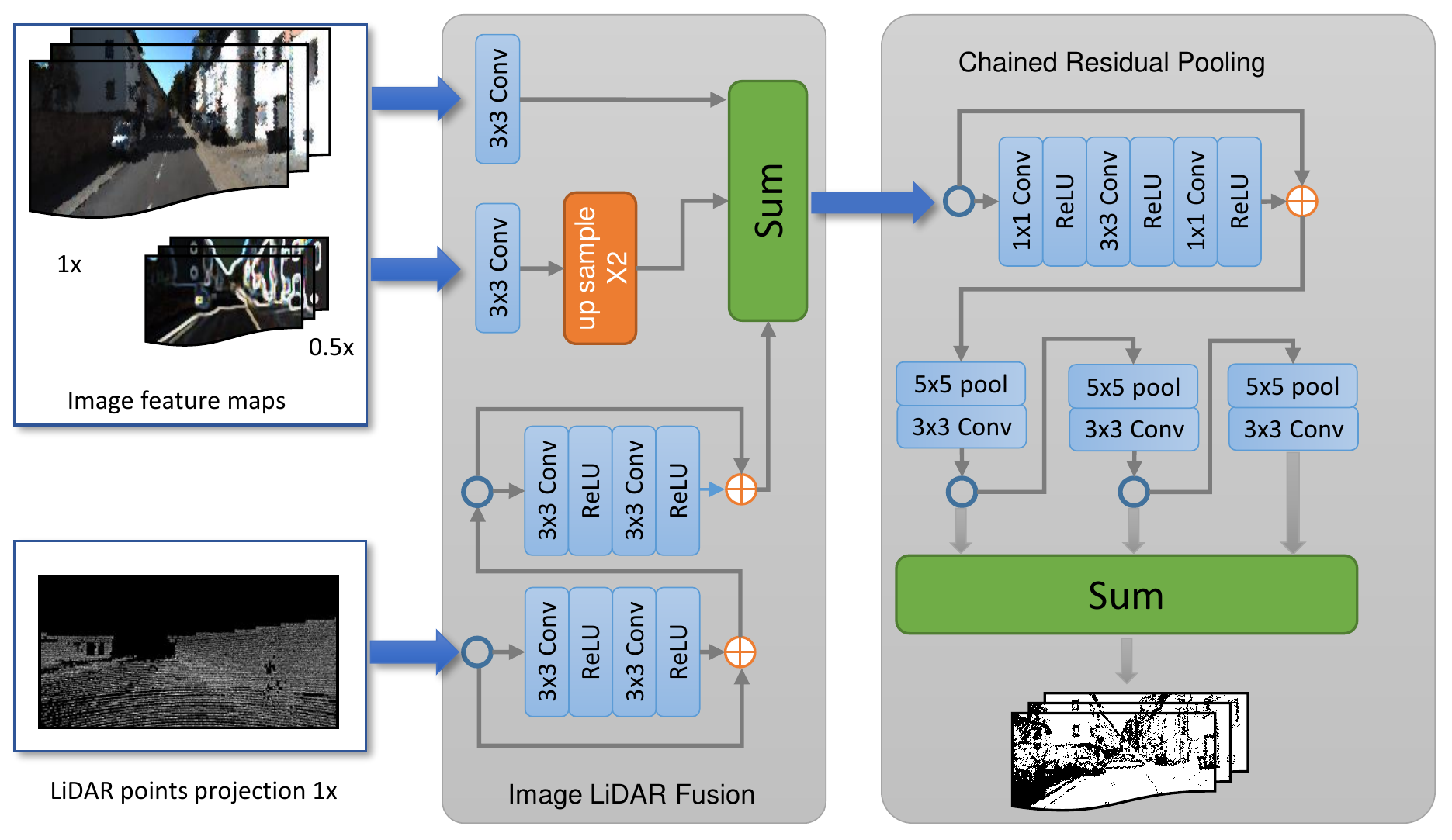}
	\caption{The Refine Fusion Unit(RFU). RFU is designed to fuse encoded image and a LiDAR points projection. Left side of this figure is input and rest of blocks is image LiDAR fusion module and a modified CRP module from RefineNet. Compared with RefineNet\cite{refinenet}, the RCU module has been removed and fusion structure is embedded. Note that pictures in this figure is just for illustration.}
	\label{fig:refine_fusion_unit}
\end{figure}

Fusing image and LiDAR data is usually understood as a sort of synergetic feature extracting problem in many previous researches. Generally, there are three ways of image-LiDAR fusion: early fusion, late fusion and stage-wise fusion. Early fusion preproccesses image and LiDAR data to create a high dimension object, then feed it to an encoder. It is quite easy to implement but not always make sense due to the unbalanced input, and at the same time, well pretrained model is hardly useful. Contrary to the early fusion, late fusion uses a group of detached encoders for each multiple input sources and finally join them each other. In this way, making use of pretrained model for each branch is possible, however, developers is required to manually adjust the fusion stages, and it is difficult to reduce the network size. Naturally, a stage-wise fusion were set forth. It performs a fusion procedure at each stage of network(usually each end of scale), then pass the fused block to next stage.  Stage-wise fusion forces the image and LiDAR information fuse with each other at each stage, which fixed the problems in both early and late fusion, but pretrained model is still useless for any input sources. The details of those three ways of fusion is illustrated in Figure. \ref{fig:three_fusion_mode}

Synergetic feature extracting in encoder is an intuitive idea but still exists some drawbacks. As we all know, CNN extracts features automatically, however, do not guarantee equilibrium between LiDAR and image usage during training process. From another point of view, we need to create massive training data to help the encoder to find out the appropriate combination of image and LiDAR features, so that features form both input sources are effectively used. Unfortunately, training data contains both color image and LiDAR point cloud with careful calibration is scarce currently, thus, over fitting is inevitable. Motivated by above discussion, we find out a practical way to avoid above issues, in which synergetic feature extracting is avoid and at the same time loading pretrained model from large scale image dataset in stead of training from scratch becomes possible.

Taken all above into consideration, we fuse LiDAR in decoding stage instead of blending it with images in encoding stage in our proposed network(see decoder in dashed bounding box, Figure. \ref{fig:architecture}(a)). In our network, LiDAR features are extracted and fused in $RFU$ just for refine the score maps. At the same time, we can use the pretrained model for encoder as a fine-tune way. The detailed structure is depicted in Figure. \ref{fig:refine_fusion_unit}

\subsection{Lidar Map generation via multi-scale reprojection}
To align the image content with LiDAR point cloud, we need generate the so called LiDAR image or LiDAR map. The first step of LiDAR image generation is 3D point projection. The projection need a set of calibration parameters between camera and LiDAR device. We assume the sensors are well calibrated in advance so that the projection matrix, including rotation and translation parameters, is already known. In practise, each frame of LiDAR data usually consist of more then 100K 3D points with three location parameters and a intensity value. Only part of those points can be projected onto the RGB image plane. Let's denote the LiDAR point clouds as $\mathbf{p}_l=(x, y, z, 1)^\mathrm{T}$ and the projection results on RGB image plane as $\mathbf{p}_i=(u, v, 1)^\mathrm{T}$. The rotation matrix is $\mathbf{R} \in \mathbb{SO}(3)$ and the translation matrix is $\mathbf{t}\in \mathbb{R}^{3\times1}$. The intrinsic parameters of camera is $\mathbf{K}\in\mathbb{R}^{3\times3}$. Then the projection can be formulated as follows:
\begin{equation}
\centering
\mathbf{p}_{img} = \mathbf{K}\cdot\mathbf{T}\cdot\mathbf{p}_{lidar} \quad, \quad \mathbf{T} = {
\left[ \begin{array}{cc} 
    \mathbf{R} & \mathbf{t} \\
    \mathbf{0} & 1
\end{array}
\right]}
\end{equation}
The projected LiDAR points is too sparse in image. Usually, we need interpolate the sparse LiDAR images to a dense ones by bilateral filter\cite{bf}, then pooling operation will generate multi-scale LiDAR maps. But we don't think interpolation is essential in this task. We introduce a scale factor $\lambda_i$ to help us reproject LiDAR images at each scale of images by pseudo intrinsic parameters $\mathcal{K} = \{\lambda_i\cdot K | \lambda_i \in \frac{1}{4}, \frac{1}{8}, \frac{1}{16}, \frac{1}{32}\}$. So the multi-scale reprejection formula is:
\begin{equation}
\centering
\mathbf{p}_{scale} = \mathcal{K}_{scale}\cdot\mathbf{T}\cdot\mathbf{p}_{img}
\end{equation}

Using LiDAR image in decoder is quite different with one in encoder. In decoder, the work flow is from small scale to large scale. Due to the projection property, densely LiDAR image is fused with pixels containing strong semantic description while. Our reprojection strategy preserved the accurate geometric information at each scale compared with arbitrary down-sampling.

\section{Experiments}
\subsection{KITTI Dataset and Data Augmentation}
We use KITTI ROAD dataset\cite{KITTI} to evaluate the performance our proposed method. There are totally 579 frames of color images and Lidar point clouds in this dataset, and their corresponding calibrations are available too. In the dataset, 289 frames of which are used as training data and the others are testing data. All the training and testing data are divided into 3 categories: UM (urban marked), UMM (urban multiple marked lanes) and UU (urban unmarked). 

Since pictures in KITTI ROAD dataset is not the same size, we performed a preprocessing step referred to \cite{yao2016icme,YYZ2017}. We pre-processed the pictures and resize color images and ground truth images to 384 by 1248, at the same time, random horizontal flips are applied with probability of 0.5. Before the images are fed to network, mean value of each channel should be subtracted.

\subsection{Implementation Details}
In the experiments, the batch size of training data is set to 4. The reprojected LiDAR point maps at 3 different scales are set to be 24*78, 48*156 and 96*312. In the first RFU, the size of input color image score maps are 24*78, and the size of input LiDAR point map is 24*78 too. After that, just like the first unit, the sizes of input image scores maps of the second unit and third unit is 48*156 and 96*312, and the sizes of input LiDAR point maps in those two units are 48*156 and 96*312.

In the begining, the learning rate of training is set to be 5e-4 for encoder and 5e-3 for decoder, and the momentum and weight decay for both encoder and decoder is 0.9and 1e-5. The total epoch number is set to 2000. The criterion is SGD and loss function is cross entropy. Before training, ResNet50 or ResNet101 should load ImageNet pretrained model. We discard trainable transpose convolutional layers for up-sampling, and use a bilinear interpolation instead. decoding procedure stop at $\frac{1}{4}$ of the original image size, and a 4x interpolation is concatenated to recover the prediction size.

\subsection{Data Fusion at Different Scales}

To verify our proposed refined fusion unit, we trained several different networks on KITTI road dataset. Specifically, since KITTI benchmark doesn't provide the ground truth of testing data, we separate the training dataset into 2 sub-datasets for training and validation. The training data contains 240 frames in training dataset, and the rest of them is validation data. After that, the training dataset is used to train various networks. They are all based on our proposed encoder-decoder architecture. The differences of them are the number of refined fusion unit they used. The  networks use 1, 2 and 3 RFU in multiple scales. We verify their performance on the validation dataset, and 2 major metrics(IoU and Accuracy).  Table \ref{tab:Table1} shows all the metrics under different configuration using ResNet-50 and ResNet-101 independently. The formulas of each metric are listed in the last section.

\begin{table}[H]
\caption{Performances of different networks on random validation data}
\centering
\begin{tabular}{ccccc}
\hline
	& ResNet-50 + 1 RFU	& ResNet-50 + 2 RFU	& ResNet-50 + 3	RFU\\
\hline
IoU                 & 94.03\%			& 95.75\%			& 96.12\%\\
Accuracy			& 95.41\%			& 96.39\%			& 96.57\%\\
\hline
 	& ResNet-101 + 1 RFU	& ResNet-101 + 2 RFU	& ResNet-101 + 3	RFU\\
\hline
IoU                 & 94.74\%			& 96.06\%			& 96.27\%\\
Accuracy			& 95.89\%			& 96.50\%			& 96.85\%\\
\hline
\end{tabular}
\label{tab:Table1}
\end{table}

From this table we can find out that the fusion of different kinds of data in multiple layers can improve the road detection result effectively. More RFU we use, the better performance the net work achieved. However, The results of our network with 2 RFUs and 3 RFUs are almost the same, indicating that the data fusion in the largest scale is not very helpful. Hence, our method only contains 3 RFUs.

\subsection{Performance on KITTI Road Dataset}
At last, we use all the 289 training frames to train our method and segment the road areas of testing frames. The results are submitted to the KITTI Benchmark Website Server. A set of metrics in bird's eye view (BEV) images are used for evaluation. They are maximum F1-measure (MaxF), average precision (AP), precision (PRE), recall (REC), false positive rate (FPR) and false negative rate (FNR).

Table \ref{tab:Table2} shows the results of our method in 3 categories and urban dataset. 
From these tables we can see, our method performs better on UM and UMM testing images, compared with UU testing images. This is because the road areas  follow more obvious spatial patterns in UM and UMM testing images. To be more specific, in those scenes, there are sidewalks or fences above the ground, separating the road and non road areas and providing much sharper edges in LIDAR point clouds.

\begin{table}[H]
\caption{Performances of our method on 3 categories of KITTI ROAD benchmark}
\centering
\begin{tabular}{ccccccc}
\hline
\textbf{Benchmark}	& \textbf{MaxF}	& \textbf{AP}	& \textbf{PRE}	& \textbf{REC}\\
\hline
UM ROAD				& 93.29\%		& 91.11\% 		& 93.53\% 		& 92.49\%\\
UMM ROAD    		& 95.05\% 		& 94.01\%		& 95.47\% 		& 95.02\%\\
UU ROAD				& 92.21\% 		& 91.62\% 		& 93.25\% 		& 94.20\%\\
URBAN ROAD			& 93.98\% 		& 92.23\% 		& 94.06\% 		& 93.90\%\\
\hline
\end{tabular}
\label{tab:Table2}
\end{table}

The results of some recently submitted real-name methods and ours are shown in Table \ref{tab:Table2}. They are DEEP-DIG\cite{DEEPDIG2017}, Up-Conv-Poly\cite{UPCONVPOLY}, HybridCRF\cite{Xiao2018} and MixedCRF\cite{Han2016}. The first two methods are deep learning based road segmentation methods, and the other two are not deep learning methods. Besides, the first two methods only use images to train the networks, while the last two methods use the fusion data of images and LIDAR point clouds. This table shows our method gain the best performance among all the methods. Compared with the results of DEEP-DIG and Up-Conv-Poly, the result of our method prove that by fusing Lidar data with images, we can improve the road segmentation ability of deep learning based methods significantly. Our method has obviously better results than HybridCRF and MixedCRF, and this shows that fusing data in deep learning framework achieves a remarkable improvement.

\begin{figure}[h]
	\centering
	\includegraphics[width=\textwidth]{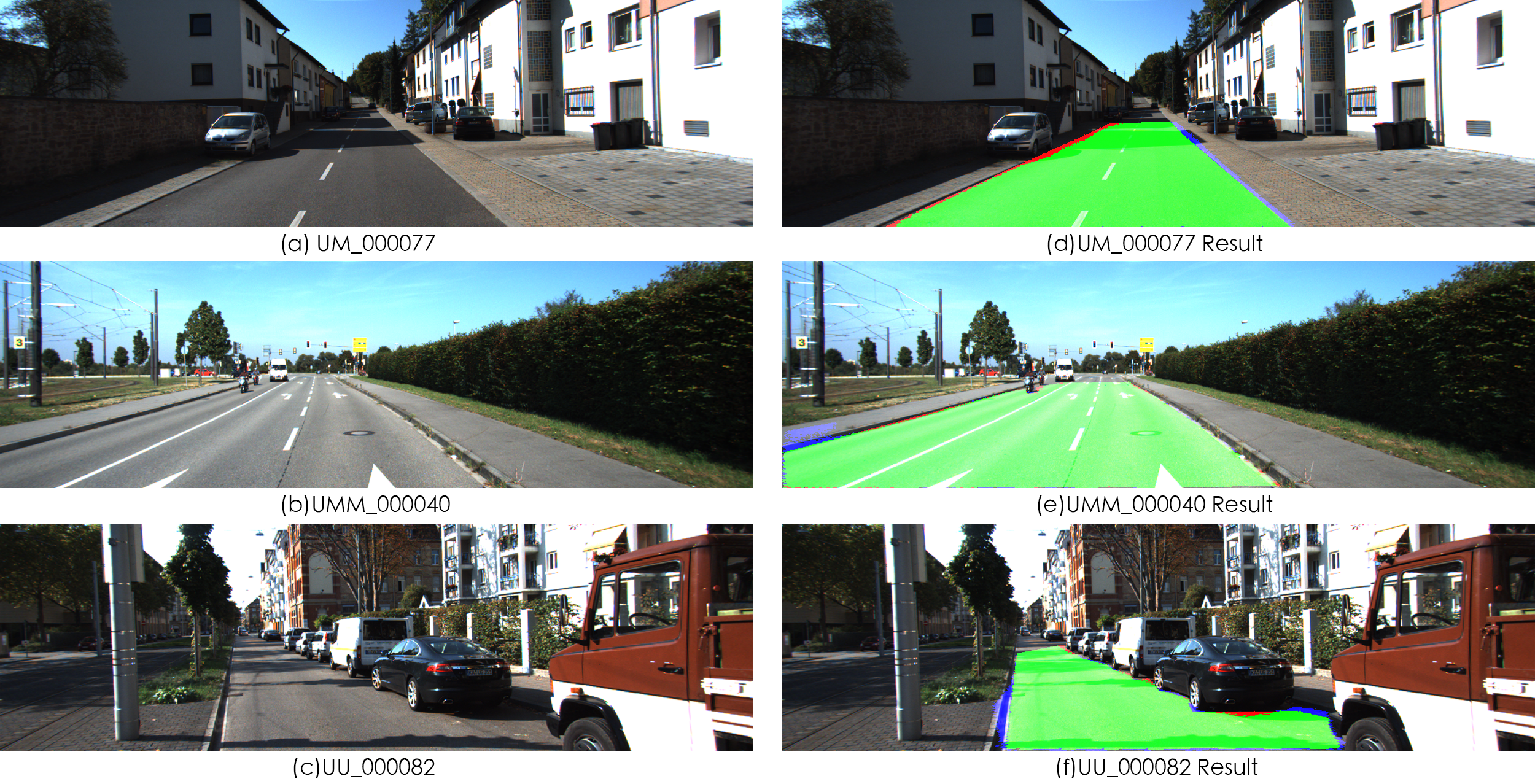}
	\caption{The results on KITTI ROAD benchmark. Figure (a), (b) and (c) are testing images, and Figure (d), (e) and (f) are the road segmentation results on those images by our method.}
	\label{fig:final_result}
\end{figure}

Figure \ref{fig:final_result} shows our final results on the KITTI ROAD benchmark in the perspective images. In the images, red areas denote false negatives, blue areas correspond to false positives and green area represent true positives. This Figure shows that our method can segment road areas in all the 3 categories effectively.

\begin{table}[H]
\caption{Performances of different methods on KITTI ROAD benchmark}
\centering
\begin{tabular}{cccccccc}
\hline
\textbf{method}	& \textbf{MaxF}	& \textbf{AP}	& \textbf{PRE}	& \textbf{REC}\\
\hline
DEEP-DIG\cite{DEEPDIG2017}		& 93.98\%		& 93.65\%		& 94.26\%		& 93.69\%\\
Up-Conv-Poly\cite{UPCONVPOLY}	& 93.83\%		& 90.47\%		& 94.00\%		& 93.67\%\\
HybridCRF\cite{Xiao2018}		& 90.81\%		& 86.01\%		& 91.05\%		& 90.57\%\\
MixedCRF\cite{Han2016}			& 90.59\%		& 84.24\%		& 89.11\%		& 92.13\%\\
Our Method						& 93.98\%		& 92.23\%		& 94.06\%		& 93.90\%\\
\hline
\end{tabular}
\label{tab:Table3}
\end{table}

\section{Discussion}
Although we have great road segmentation performances, our method is a bit overfitting to the urban road scenes, facing the lack of annotated data in other environments. And this problem leads to the relying on a pre-trained model, just like we address in the Section I. More data should be annotated so that the  network with data fusion in deeper layers can be trained in the future.

\section{Conclusions}

This paper propose a novel structure to fuse image and LiDAR point cloud in an end-to-end semantic segmentation network. The fusion is performed at decoder stage. We exploit the multi-scale LiDAR maps which generated from LIDAR point clouds by using pyramid projection method. to fuse with the image features in different layers. Additionally, we adapted the multi-path refinement network with our fusion strategy and improve the road segmentation results compared with transpose convolution with skip layers. Our approach has been tested on KITTI ROAD dataset and have a competitive performance.

\section{Notes}
In this section, we list some details for the notation and indicators mentioned above. In following equations, $T$ is short for TRUE, $F$ is short for FALSE, $P$ is short for POSITIVE and $N$ is short for NEGATIVE. $PRE$ is short for $precision$, $REC$ is short for$recall$, $MaxF$ is $F\textrm{-}measure$. The definition is shown as follows:
\begin{eqnarray*}
&& PRE = \frac{TP}{TP+FP} \\
&& REC = \frac{TP}{TP+FN} \\
&& MaxF = \frac{2\times PRE \times REC}{PRE+REC} \\
&& Accuracy = \frac{TP+TN}{TP+TN+FP+FN}
\end{eqnarray*}

\end{document}